\documentclass[lettersize,journal]{IEEEtran}
\usepackage{amsmath,amsfonts}
\usepackage{array}
\usepackage[caption=false,font=normalsize,labelfont=sf,textfont=sf]{subfig}
\usepackage{textcomp}
\usepackage{stfloats}
\usepackage{url}
\usepackage{mathtools}
\usepackage{verbatim}
\usepackage{multirow}
\usepackage{graphicx}
\usepackage{booktabs}
\usepackage{colortbl}
\usepackage{xcolor}
\usepackage{hyperref}
\definecolor{V}{RGB}{21,137,139}
\definecolor{X}{RGB}{234,120,60}
\usepackage{cite}
\usepackage{algorithm}
\usepackage{algpseudocode}
\usepackage[algo2e]{algorithm2e}
\usepackage[utf8]{inputenc}
\begin{document}

\title{
Beyond Low-Rank: 
Low-Rank Sparse Prompting via  Spiking Neural Network and Prompt Factorization
}

\author{Yumiao Zhao, Bo Jiang*, Beibei Wang, Xixi Wan, Xiao Wang, Jin Tang 
\thanks{

The authors are with the Information Materials and Intelligent
Sensing Laboratory of Anhui Province, Anhui Provincial Key Laboratory of
Multimodal Cognitive Computation, School of Computer Science and Technology, Anhui University

*Corresponding author

}
\thanks{Manuscript received April 19, 2021; revised August 16, 2021.}}

\markboth{Journal of \LaTeX\ Class Files,~Vol.~14, No.~8, August~2021}%
{Shell \MakeLowercase{\textit{et al.}}: A Sample Article Using IEEEtran.cls for IEEE Journals}

\IEEEpubid{0000--0000/00\$00.00~\copyright~2021 IEEE}
\maketitle
\begin{abstract}
Visual Prompting (VP) has emerged as an efficient paradigm for adapting large-scale pre-trained vision models to downstream tasks by incorporating learnable prompts at the input level. However, existing VP methods typically employ dense pixel-level prompts, which often suffer from redundant perturbations, limited generalization and energy inefficiency. 
To overcome these limitations, 
we propose to integrate brain-inspired spiking learning 
into visual prompt learning tasks.  
As we know that spiking neuron can perform inexpensive information processing by
transmitting the input data into discrete spike trains and return
sparse outputs.
Inspired by this, we propose \textbf{Lo}w-\textbf{R}ank visual \textbf{S}pike \textbf{P}rompting (LoRSP), a novel framework that learns dynamic low-rank sparse visual prompts naturally via a Spiking neuron learning mechanism. The core idea of LoRSP is to exploit the  brain-inspired  sparse firing mechanism of spiking neurons to generate pixel-level sparse prompt for each instance. 
To be specific, we first construct a series of prompt factors via low-rank factorization to capture distinct prompt subspaces. 
These prompt factors are then fed into an SNN architecture, which performs the integrate-and-fire process to emit spikes. 
As a result, our LoRSP generates a \emph{sparse} visual prompt while maintaining the low-rank constraint. 
This design enables instance-specific selective prompting, leading to more compact and robust adaptation across diverse downstream tasks. 
Extensive experiments on five heterogeneous vision backbones and multiple benchmarks demonstrate that LoRSP achieves competitive performance while requiring fewer tunable parameters compared to existing VP methods.
\end{abstract}

\begin{IEEEkeywords}
Visual Prompting, Spiking Neural Networks,  Sparse Prompts, Low-Rank Matrix Factorization.
\end{IEEEkeywords}

\section{Introduction}
\IEEEPARstart {L}{arge-scale} vision architectures, such as Vision Transformer (ViT)~\cite{dosovitskiy2020image} and Swin Transformer~\cite{liu2021swin}, have emerged as foundational backbones for visual representation learning by pre-training on massive datasets. These models exhibit strong generalization across diverse application scenarios, such as image classification~\cite{gao2024clip,silva2024closer}, semantic segmentation~\cite{bi2024prompt,li2024relationship}, and medical imaging~\cite{wei2025more,yin2025ddfp}. To efficiently adapt large-scale pre-trained vision models to downstream tasks, the “pretrain-then-finetune” paradigm is widely used. Conventional full fine-tuning (FT) updates all model parameters for downstream tasks, requiring high computational costs and being prone to catastrophic forgetting. Inspired by the success of prompting in large language models (LLMs)~\cite{liu2023pre,liu2022few}, Parameter-Efficient Fine-Tuning (PEFT) has gained attention as a promising paradigm for adapting frozen vision backbones with minimal trainable parameters. There are two primary paradigms in parameter-efficient fine-tuning (PEFT) for visual models: Visual Prompting (VP)~\cite{tsao2023autovp,jin2025lor,chen2023understanding} and Visual Prompt Tuning (VPT)~\cite{jia2022visual,liu2024insvp,ren2025vpt}. 
VPT adapts pre-trained models by inserting learnable prompt tokens into intermediate layers, whereas VP operates directly in the input pixel space. 
By modifying the input and keeping the backbone frozen, VP is architecture-agnostic and easily transfers diverse vision backbones across Convolutional Neural Networks (CNNs) and vision transformers with minimal additional parameters. 
\begin{figure}[t]
  \centering
  \includegraphics[width=1\linewidth]{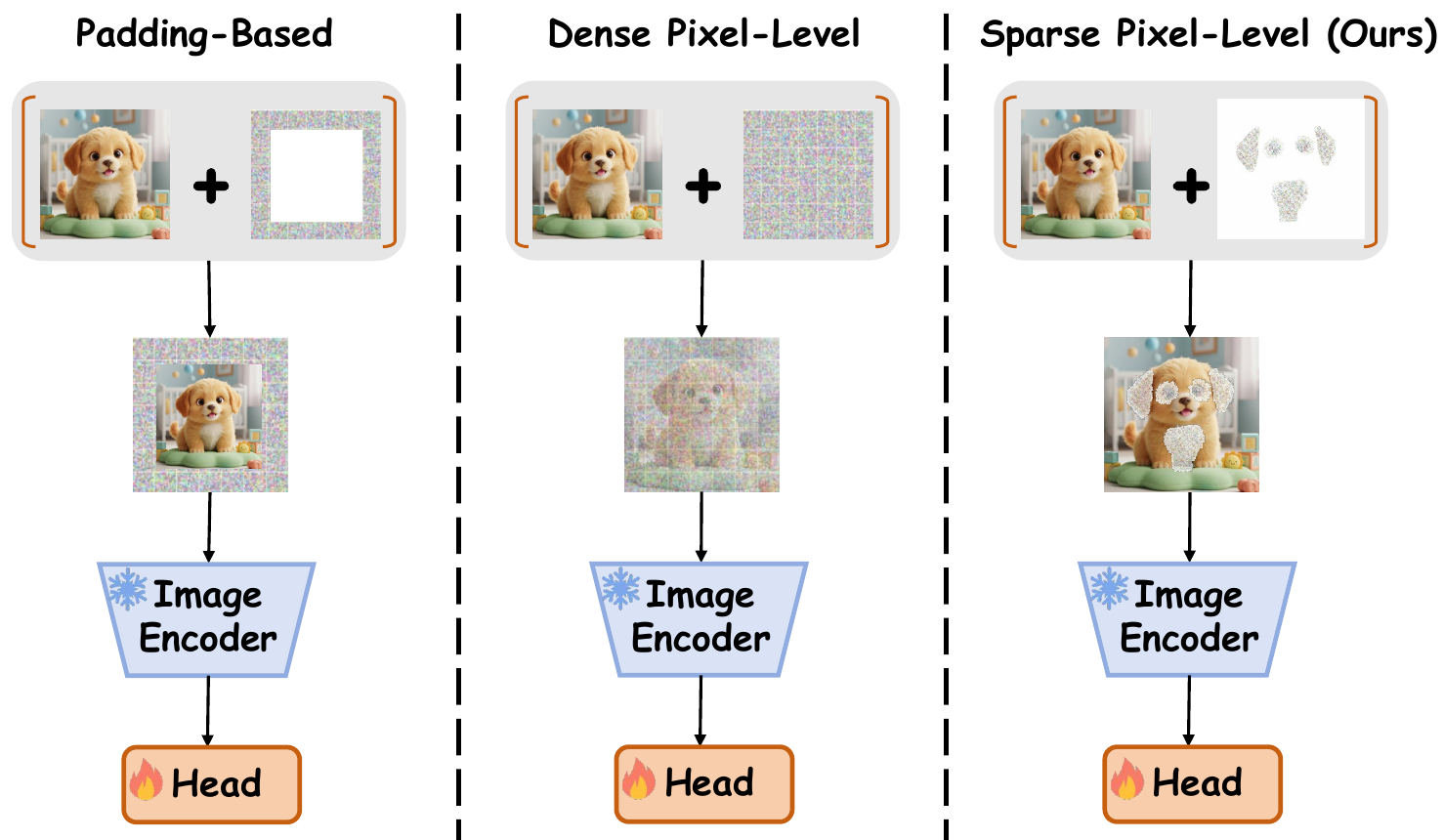}
  \caption{Illustration of various Visual Prompting (VP) paradigms. (Left) Padding-Based: This method pads learnable prompts around the resized input image. (Middle) Dense Pixel-Level: This approach generates a learnable prompt for each pixel of the input image. (Right) Sparse Pixel-Level (Ours): Our proposed LoRSP learns instance-specific sparse visual prompts that focus on task-relevant information while filtering out redundant prompt responses.}
\label{motivation}
\end{figure}

\IEEEpubidadjcol
Existing VP methods can be broadly categorized into two types: padding-based prompting and dense pixel-level prompting. Padding-based methods pad learnable prompts around the resized input image, as illustrated in Fig.~\ref{motivation} (Left). Representative approaches such as AutoVP~\cite{tsao2023autovp} automatically search for the size and shape of padding-based prompts around the resized image. However, padding-based prompting primarily affects peripheral regions of the image while preserving the central content, limiting its ability to refine task-relevant visual content for downstream adaptation.
In contrast, dense pixel-level prompting assigns learnable prompts to every pixel of the input image, as shown in Fig.~\ref{motivation} (Middle).
For example, LoR-VP~\cite{jin2025lor} generates dense pixel-level prompts via a low-rank matrix multiplication, and ILM-VP~\cite{chen2023understanding} learns a shared prompt map in the pixel space. However, these methods are typically instance-invariant (shared across images), which limits their ability to adapt to diverse image samples. Furthermore, applying dense prompts across the entire image may introduce redundant perturbations and disrupt the underlying spatial structure, leading to suboptimal representations.

Given these limitations, a natural question arises: \textbf{Can we design a compact and instance-specific prompt generator that selects \emph{where} to prompt and \emph{how strongly} to prompt?} A straightforward way is to impose sparse constraints on dense prompts. Conventional sparsification strategies include Top-$k$ selection~\cite{jayakumar2020top,han2023e2vpt}, random dropout~\cite{srivastava2014dropout,}, and fixed-threshold selection~\cite{tun2020network,yang2024exploring}. 
Top-$k$ selection retains a fixed number of responses and random dropout masks responses stochastically. Fixed-threshold selection suppresses responses according to a predefined threshold. 
However, in these methods, the sparse constraint is enforced independently of prompt learning, leading to a suboptimal solution. Moreover, they fail to exploit contextual information across prompt maps for adaptive sparse selection.

It is well known that Spiking Neural Network (SNN)~\cite{kim2023sharing,zhou2024direct,jiang2026prompting} can perform information processing across sequential data and naturally generate sparse outputs. 
Inspired by this, we propose \textbf{Lo}w-\textbf{R}ank visual \textbf{S}pike \textbf{P}rompting (LoRSP), which leverages \textbf{low-rank prompt factorization} and \textbf{SNN} to generate pixel-level low-rank sparse prompts for each instance. 
Specifically, we first construct a series of low-rank prompt factors $\{\mathbf{B_1A}, \mathbf{B_2A}, \dots, \mathbf{B_{k}A}\}$ via matrix factorization to capture distinct semantic subspaces where the low-rank basis $\mathbf{A}$ is shared across different instances, 
Then, we feed these prompt factors into an SNN model, which performs the integrate-and-fire~\cite{maass1997networks,su2026pts} process to emit spikes and outputs a sparse solution. 
As a result, our LoRSP generates a sparse visual prompt while preserving the low-rank constraint. 
This design enables instance-specific selective prompting, leading to compact and robust adaptation across various downstream tasks.  
Note that, compared to traditional Top-$k$ and fixed-threshold selection strategies
~\cite{jayakumar2020top,tun2020network}, our LoRSP naturally returns a sparse prompt solution and achieves better optimality by exploiting contextual information across different prompt factors. 
We note that 
Spiking Neural Networks (SNNs) have previously been  employed for designing adapters and prompts~\cite{ji2025snnptrack,su2026pts}. Different from these works, we leverage SNN with prompt factorization to learn \textbf{sparse, low-rank} instance prompts in a unified framework.

Overall, the primary contributions of this paper are summarized as follows:
\begin{itemize}
\item 
We propose integrating a SNN architecture into prompt learning to enable compact and robust adaptation of pre-trained models across diverse downstream tasks. To the best of our knowledge, this paper is the first work to leverage SNNs for \emph{sparse visual prompt} learning. 

\item 
We propose a novel \textbf{Lo}w-\textbf{R}ank visual \textbf{S}pike \textbf{P}rompting (LoRSP) approach that integrates low-rank and sparse constraints into a parameter-efficient framework, generating instance-adaptive pixel-level sparse prompts for adapting pre-trained vision models. 

\item Extensive experiments across five heterogeneous vision backbones and multiple benchmarks demonstrate that LoRSP achieves competitive performance while requiring fewer tunable parameters. 
\end{itemize}

\begin{figure*}[htbp]
  \centering
  \includegraphics[width=0.95\linewidth]{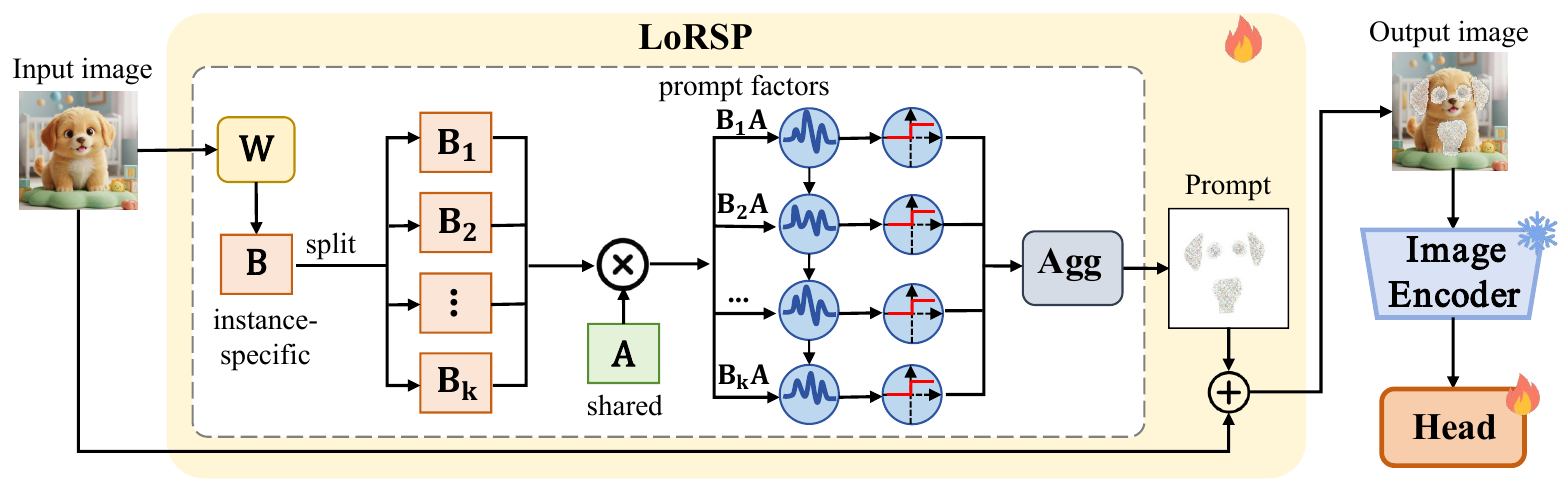}
  \caption{ The overall framework of the proposed \textbf{Lo}w-\textbf{R}ank visual \textbf{S}pike \textbf{P}rompting (LoRSP). LoRSP first applies low-rank factorizations to generate group-wise prompt factors. These factors are then fed into the SNN module to generate pixel-level sparse prompts for each instance.
  } 
\label{network}
\end{figure*}

\section{Related Work}
\label{relatedwork}

\subsection{Visual prompting}
Visual prompting (VP)~\cite{liu2024insvp,wu2022unleashing,cai2024bayesian,cai2025attribute} introduces learnable prompts in the input space as a parameter-efficient strategy for adapting pre-trained models to downstream tasks, which is particularly effective in low-data regimes. A typical VP framework consists of two trainable modules: an input transformation module that incorporates prompts into the input, and an output transformation module that maps the model outputs to task-specific predictions. Input-side VP mainly includes padding-based and dense pixel-level prompting. The padding-based prompting methods introduce learnable prompts around the image. Dense pixel-level prompting assigns learnable prompts to every pixel of the input image. Some methods learn shared pixel prompts, while others design generators to produce instance-adaptive prompts in the input space. For example, OT-VP~\cite{zhang2025ot} aligns source and target distributions with an Optimal Transport objective to learn a shared prompt. Ddfp~\cite{yin2025ddfp} uses a frequency prompt to shift target images toward the source domain. Blackvip~\cite{oh2023blackvip} trains a coordinator to produce input-specific prompts. For output transformation, the goal is to map source model outputs to target task labels without updating the full model. For instance, Tsai et al.~\cite{tsai2020transfer} introduce frequency-based label mapping (FLM), which establishes label mapping by computing matching frequencies between source and target labels. ILM-VP~\cite{chen2023understanding} proposes iterative label mapping (ILM) to progressively refine label alignment. Yang et al.~\cite{yang2023prompt} propose a semantic label mapping strategy that aligns source and target classes based on their semantic similarity.

\subsection{Spiking neural network}
Spiking neural networks (SNNs)~\cite{guo2025spiking,yao2023spike,wei2024event} are inspired by biological neural systems and encode information as discrete spikes. Unlike conventional Artificial Neural Networks (ANN) that rely on dense continuous activations, SNNs naturally produce temporally and spatially sparse representations, thereby enabling more efficient computation. The Leaky Integrate-and-Fire (LIF) model~\cite{maass1997networks} is a representative and lightweight neuron model for simulating neuronal dynamics. SNNs have been extensively studied for computer vision tasks~\cite{cao2015spiking,cao2024spiking,luo2024integer,guo2022recdis}, including image classification, object detection, and video understanding. For example, Spiking-yolo~\cite{kim2020spiking} utilizes a channel-wise activation normalization strategy to refine the mapping between ANNs and SNNs, thereby realizing real-time object detection while significantly reducing power consumption. Spiking-PointNet~\cite{ren2023spiking} adopts SNNs for 3D point cloud processing, leveraging the inherent sparsity of spiking neurons to significantly alleviate the storage and computational burden associated with large-scale point cloud data. Yang et al.~\cite{yang2025temporal} introduce a temporal-guided SNN framework for event-based human action recognition, thereby overcoming the limitations of conventional SNNs in processing long-range sequences while maintaining high energy efficiency. Zhou et al.~\cite{zhou2026dynamic} incorporate dynamic synaptic weight adaptation into SNNs for complex obstacle avoidance and target tracking, thereby empowering the model with self-regulation capabilities to adapt to dynamic scene changes without manual fine-tuning. 

\section{Approach}
\subsection{Preliminaries}
\label{Preliminaries}
Spiking Neural Networks (SNNs) adopt spiking neurons as the basic computational unit to achieve spiking neural dynamics. The spiking neuron first integrates spatial and temporal information from input signals, accumulating them into a membrane potential over time. Then, when the membrane potential reaches a predefined threshold, the neuron emits a spike. A widely used model for implementing such spiking dynamics is the classical Leaky Integrate-and-Fire (LIF) neuron~\cite{maass1997networks}. Specifically, given the input current $\mathbf{X}_{t}$ at time step $t$, the above learning process in the LIF model can generally be formulated as follows:
\begin{align}
\mathbf{U}_{t}&=\mathbf{U}_{t-1} +\frac{1}{\beta}(\mathbf{X}_{t}-(\mathbf{U}_{t-1}-V_{\text{rst}}))\label{eq:prelim_1} \\
\mathbf{S}_{t} &= \Theta\left(\mathbf{U}_{t}- V_{\mathrm{th}}\right)\label{eq:prelim_2} \\
\mathbf{U}_{t} &=\begin{cases}
\mathbf{S}_{t}V_{\text{rst}}+(1-\mathbf{S}_{t})\mathbf{U}_{t}, & \text{fixed-reset,}\\
\mathbf{U}_{t}-V_{\mathrm{th}}\mathbf{S}_{t}, & \text{subtraction-reset.}
\end{cases}\label{eq:prelim3}
\end{align}
where $\mathbf{U}_{t}$ represent the membrane potential. $V_{\mathrm{th}}$ represents the firing threshold and $V_{\text{rst}}$  is the rest parameter. Specifically, Eq.~(\ref{eq:prelim_1}) defines the leaky integration process, where the neuron aggregates the current input $\mathbf{X}_{t}$ while exhibiting an exponential decay controlled by the leak factor $\beta$. Eq.~(\ref{eq:prelim_2}) describes the spiking neuron fire mechanism, where a binary spike $\mathbf{S}_{t}$ is triggered via the Heaviside step operator $\Theta(\cdot)$, i.e., $\Theta(x)=1$ if $x \ge 0$ and $\Theta(x)=0$ otherwise. Eq.~(\ref{eq:prelim3}) executes membrane potential for the next time step. 

%

\subsection{Low-rank visual spike prompting}
Existing Visual Prompting (VP) methods typically employ dense pixel-wise prompts across the entire input image, which may introduce redundant perturbations and thus lead to suboptimal generalization performance. To overcome this limitation, a natural solution is to learn sparse and selective prompts for VP tuning. However, conventional sparsification strategies~\cite{jayakumar2020top,srivastava2014dropout,tun2020network} are designed independently of prompt learning and do not fully exploit contextual information for adaptive sparse prompt learning. Therefore, we propose \textbf{Lo}w-\textbf{R}ank visual \textbf{S}pike \textbf{P}rompting (LoRSP), which leverages an SNN to adaptively learn instance-specific sparse prompts by determining \emph{where} to prompt and \emph{how strongly} to prompt.
As illustrated in Fig.~\ref{network}, LoRSP first leverages low-rank matrix factorizations to generate group-wise prompt factors, capturing distinct semantic subspaces. Then, these prompt factors are viewed as the spiking sequence and fed into the SNN module to learn a sparse visual prompt for the input image.

\subsubsection{Group-wise prompt factors}
Given an input image $\mathbf{I}$, we first resize it to a resolution of $c\times L\times L$, which is denoted $\hat{\mathbf{I}}$. Prior LoRA-based prompt learning works~\cite{jin2025lor,hu2022lora} realize downstream adaptation using a shared low-rank decomposition $\mathbf{B}\mathbf{A}$, where the matrices $\mathbf{A}$ and $\mathbf{B}$ are shared for all images. To better adapt to each input image while preserving the benefits of the low-rank constraint, we learn a shared low-rank basis $\mathbf{A}$ to capture task-shared knowledge, while using instance-specific matrices $\{\mathbf{B}_1,\mathbf{B}_2,\cdots,\mathbf{B}_k\}$ to encode instance-specific cues. Through a series of low-rank matrix factorizations, we construct group-wise prompt factors $\mathbf{F}=\{\mathbf{F}_1, \mathbf{F}_2\cdots \mathbf{F}_t\}$ to capture distinct semantic subspaces as:
\begin{align}
&\{\mathbf{B}_1, \mathbf{B}_2, \dots, \mathbf{B}_k\}=\mathrm{Split}(\hat{\mathbf{I}}\mathbf{W}),\\
&\mathbf{F}_{t}=\mathbf{B}_t \mathbf{A}, \;\text{where}\; t={1,\cdots,k}
\end{align}
where $\mathbf{W}\in \mathbb{R}^{L\times(k\times r)}$ is a trainable spatial projection matrix. The dimensions of $\mathbf{A}$ and $\mathbf{B}_t$ are $c\times r\times L$ and $c\times L\times r$, respectively. Here, $c$ denotes the number of channels for the input image, and $r\ll L$ is the decomposition rank. The $\mathrm{Split}(\cdot)$ operation is used to partition the output into $k$ groups along the last dimension.

\subsubsection{SNN based sparse prompt learning}

Previous works~\cite{kim2023sharing,zhou2024direct} show that Spiking Neural Networks (SNNs) are well-suited for modeling sequence dependencies and generating sparse outputs. Therefore, we naturally leverage SNNs to develop our sparse prompt learning mechanism and capture contextual information among group-wise prompt factors. To be specific, we treat the prompt factors $\mathbf{F}=\{\mathbf{F}_t\}^{k}_{t=1}$ as a sequence and feed them into a SNN model to obtain instance-specific sparse visual prompts. The learning process is defined as follows:
\begin{align}
&\mathbf{U}_{t} = \mathbf{U}_{t-1} +\frac{1}{\beta}(\mathbf{F}_{t}-\mathbf{U}_{t-1}),\\
&\mathbf{S}_{t} = \Theta\left(\mathbf{U}_{t}- V_{\mathrm{th}}\right)=\begin{cases}
1, & \text{if } U_{t} \ge V_{\mathrm{th}},\\
0, & \text{otherwise.}
\end{cases},\\
&\mathbf{U}_{t} =
\mathbf{U}_{t}-V_{\mathrm{th}}\mathbf{S}_{t}.
\end{align}
where $\mathbf{F}_{t}$ represents the $t$-th prompt factor. $\beta$ represents the leak factor. $V_{\mathrm{th}}$ denotes the firing threshold. As shown in Eq.(7), Once the membrane potential exceeds the firing threshold, the neuron emits sparse spikes $S_t$. 
Then, $\mathbf{S}_{t}$ is used to achieve sparse visual prompt generation as
\begin{equation}
\mathbf{P} 
= \frac{1}{k}\sum_{t=1}^{k} (\mathbf{S}_{t}\odot \mathbf{F}_{t}),
\end{equation} 
where $\mathbf{P}$ denotes the learned sparse prompt. 
Since $\mathbf{S}_i$ is sparse and $\mathbf{F}_t$ is low-rank, the output $\mathbf{P}$ inherits these properties, thereby being low-rank and sparse simultaneously.   
Therefore, our LoRSP learns sparse low-rank instance prompts in a unified framework, thereby suppressing redundant prompt responses and enabling instance-specific selective prompting.

\subsection{Adaptation for downstream tasks}
Given a resized input image $\hat{\mathbf{I}}$ with label $i$, we add the generated sparse visual prompt $\mathbf{P}$ to obtain the prompted image $\tilde{\mathbf{I}}$. The prompted image $\tilde{\mathbf{I}}$ is fed into a frozen vision backbone $\mathcal{V}(\cdot)$ to extract visual representations. To align the backbone output with the target label space, we follow previous work~\cite{tsao2023autovp,jin2025lor} and use Linear Probing (LP) as output transformation. The model is optimized with the standard cross-entropy loss as
\begin{align}
\tilde{\mathbf{I}}&=\hat{\mathbf{I}}+\mathbf{P}, \\
\hat{\mathbf{y}} &= f\left(\mathcal{V}(\tilde{\mathbf{I}})\right),\\
\mathcal{L}_{\mathrm{CE}}(\hat{\mathbf{y}}, i) &= - \log\frac{
\exp\left( \hat{\mathbf{y}}_{i} \right)
}{
\sum_{c=1}^{C} \exp\left( \hat{\mathbf{y}}_{c} \right)
},
\end{align}
where $f(\cdot)$ denotes output transformation operation. $\hat{\mathbf{y}}$ represents the predicted logits, and $C$ denotes the number of categories.

\section{Experiments}
\label{experiment}
\subsection{Experimental setup}
\subsubsection{Datasets} 
To comprehensively evaluate the effectiveness of the proposed \textbf{Lo}w-\textbf{R}ank visual \textbf{S}pike \textbf{P}rompting (LoRSP) framework, we conduct extensive experiments across three tasks. First, we validate the effectiveness of LoRSP with various vision backbones on the Tiny-ImageNet~\cite{le2015tiny} and CIFAR-100~\cite{krizhevsky2009learning} datasets. Second, we evaluate the generalization of our method across diverse downstream visual recognition tasks spanning multiple domains, including fine-grained, remote sensing, and texture classification. We conduct experiments on Tiny-ImageNet~\cite{le2015tiny}, EuroSAT~\cite{helber2019eurosat}, OxfordPets~\cite{parkhi2012cats}, Food101~\cite{bossard2014food}, DTD~\cite{cimpoi2014describing}, Flowers102~\cite{nilsback2008automated}, CIFAR-10~\cite{krizhevsky2009learning}, and CIFAR-100~\cite{krizhevsky2009learning}. Finally, to verify the Out-Of-Distribution (OOD) robustness of LoRSP, we train the model on ImageNet-1K~\cite{deng2009imagenet} dataset and evaluate its performance on four challenging datasets, including ImageNet-V2~\cite{recht2019imagenet}, ImageNet-Sketch~\cite{wang2019learning}, ImageNet-A~\cite{hendrycks2021natural}, and ImageNet-R~\cite{hendrycks2021many}.

\subsubsection{Vision backbone}
To evaluate the robustness of LoRSP across heterogeneous model architectures, we perform experiments on five large-scale pre-trained vision backbones, including ResNet-50~\cite{he2016deep}, ViT-B/32~\cite{dosovitskiy2020image}, ViT-B/16~\cite{dosovitskiy2020image}, Swin-B~\cite{liu2021swin}, and CLIP~\cite{radford2021learning} (with ViT-B/32 vision encoder). These backbones represent diverse architectural paradigms, covering convolutional, transformer-based, and vision–language models. The pre-training datasets and corresponding model parameters for each backbone are detailed in Table~\ref{model_details}. All pre-trained weights are publicly available and are obtained from the official PyTorch Model Zoo or the Hugging Face Timm library.
\begin{table}[htbp]
  \centering
  \caption{Details of the vision backbones}
    \begin{tabular}{c|c|c}
    \toprule
    Vision Backbone & Pre-trained Dataset & Model Params \\
    \midrule
    ResNet-50~\cite{he2016deep} & ImageNet-1K & 26M \\
    ViT-B/32~\cite{dosovitskiy2020image} & ImageNet-21K, ImageNet-1K & 88M \\
    ViT-B/16~\cite{dosovitskiy2020image} & ImageNet-21K-P & 94M \\
    Swin-B~\cite{liu2021swin} & ImageNet-21K & 109M \\
    CLIP~\cite{radford2021learning}  & WebImageText & 86M \\
    \bottomrule
    \end{tabular}%
  \label{model_details}%
\end{table}%

\begin{table*}[htbp]
  \centering
  \caption{Performance comparison of LoRSP with representative visual prompting methods across five pre-trained vision backbones on Tiny-ImageNet and CIFAR-100. The ``*'' denotes results reproduced from the officially released implementations.}
   \tiny
  \resizebox{0.8\textwidth}{!}{
    \begin{tabular}{c|c|ccccc|c}
    \toprule
    \multirow{2}[4]{*}{Dataset} & \multirow{2}[4]{*}{Method} & \multicolumn{5}{c|}{Vision Backbone}          & \multirow{2}[4]{*}{Avg} \\
\cmidrule{3-7}          &       & ResNet-50 & ViT-B/32 & ViT-B/16 & Swin-B & CLIP  &  \\
    \midrule
    \multirow{6}[2]{*}{Tiny-ImageNet} & ILM-VP & 37.74 & 32.58 & 26.17 & 56.28 & 65.66 & 43.69  \\
          & *LP   & 76.52 & 83.95 & 88.86 & 86.54 & 67.93 & 80.76  \\
          & AutoVP & 77.04 & 82.43 & 81.42 & 84.81 & 67.44 & 78.63  \\
          & *EVP  & 77.04 & 85.83 & 90.03 & 84.91 & 71.34 & 81.83  \\
          & *LoR-VP & 77.25 & 85.77 & 89.78 & 88.02 & 71.58 & 82.48  \\
          & \cellcolor[rgb]{ .906,  .902,  .902}LoRSP (Ours) & \cellcolor[rgb]{ .906,  .902,  .902}\textbf{78.27} & \cellcolor[rgb]{ .906,  .902,  .902}\textbf{86.93} & \cellcolor[rgb]{ .906,  .902,  .902}\textbf{90.78} & \cellcolor[rgb]{ .906,  .902,  .902}\textbf{88.46} & \cellcolor[rgb]{ .906,  .902,  .902}\textbf{74.06} & \cellcolor[rgb]{ .906,  .902,  .902}\textbf{83.70} \\
    \midrule
    \multirow{6}[2]{*}{ CIFAR-100} & ILM-VP & 35.68 & 40.1  & 41.49 & 65.78 & 73.91 & 51.39  \\
          & *LP   & 73.18 & 86.21 & 88.90  & 88.52 & 78.15 & 82.99  \\
          & AutoVP & 72.69 & 85.96 & 88.58 & 86.83 & 77.08 & 82.23  \\
          & *EVP  & 75.52 & 88.19 & 88.66 & 89.21 & 78.89 & 84.09  \\
          & *LoR-VP & 75.06 & 88.06 & 88.97 & 89.73 & 79.19 & 84.20  \\
          & \cellcolor[rgb]{ .906,  .902,  .902}LoRSP (Ours) & \cellcolor[rgb]{ .906,  .902,  .902}\textbf{76.34} & \cellcolor[rgb]{ .906,  .902,  .902}\textbf{88.69} & \cellcolor[rgb]{ .906,  .902,  .902}\textbf{89.62} & \cellcolor[rgb]{ .906,  .902,  .902}\textbf{90.16} & \cellcolor[rgb]{ .906,  .902,  .902}\textbf{80.36} & \cellcolor[rgb]{ .906,  .902,  .902}\textbf{85.03} \\
    \bottomrule
    \end{tabular}%
  \label{tab:vis}%
  }
\end{table*}%

\begin{table*}[htbp]
  \centering
  \caption{Performance comparison of LoRSP with representative visual prompting methods across downstream classification tasks using ViT-B/32 as the pre-trained backbone.}
    \begin{tabular}{c|cccccccc|c}
    \toprule
    Method & Tiny-ImageNet & EuroSAT  & OxfordPets &  Food101 & DTD   & Flowers102 & CIFAR-10 &  CIFAR-100 & Avg \\
    \midrule
    LP    & 83.95 & 95.67 & 91.9  & 82.18 & 69.83 & 97.98 & 96.51 & 86.21 & 88.03  \\
    VPT   & 83.54 & 95.9  & 92.27 & 82.29 & 72.11 & 98.47 & 96.02 & 86.22 & 88.35  \\
    AutoVP & 82.43 & 96.25 & 92.12 & 82.86 & 70.81 & 98.42 & 95.45 & 85.96 & 88.04  \\
    LoR-VP & 85.77 & 96.25 & 92.18 & 83.51 & 72.49 & 98.58 & 97.52 & 88.06 & 89.30  \\
    \rowcolor[rgb]{ .906,  .902,  .902} LoRSP (Ours) & \textbf{86.93} & \textbf{97.09} & \textbf{93.54} & \textbf{84.36} & \textbf{76.71} & \textbf{99.19} & \textbf{97.77} & \textbf{88.69} & \textbf{90.54} \\
    \bottomrule
    \end{tabular}%
  \label{tab:tasks}%
\end{table*}%

\subsubsection{Implementation details}
Following prior works~\cite{tsao2023autovp,jin2025lor,wu2024unleashing}, the proposed LoRSP adopts Linear Probing (LP) as the output transformation for downstream adaptation. All input images are resized to $224\times 224$. We set $k=4$ and $r=4$ for downstream adaptation across various vision backbones, where $k$ denotes the number of prompt factors and $r$ represents the rank used in the low-rank factorization. The number of trainable parameters introduced by visual prompting is about 6.27K. For the out-of-distribution (OOD) robustness assessments, we set $k=2$ and $r=2$, which introduce only 2.24K trainable parameters. All experiments are trained for 20 epochs. We use Stochastic Gradient Descent (SGD)~\cite{bottou2010large} for optimization with a weight decay of $1e-4$. The momentum is set to 0.9. We use a batch size of 256 and initialize the learning rate to 0.01. The $V_{\mathrm{th}}$ and $\beta$ in LoRSP are set to 0.0 and 4.0, respectively. All experiments are conducted on a single NVIDIA RTX 3090 GPU with 24GB of memory.

\subsection{Comparison with state-of-the-art methods}
\subsubsection{Performance across vision backbones}
Table~\ref{tab:vis} reports the performance of LoRSP compared with representative visual prompting methods across five heterogeneous vision backbones on Tiny-ImageNet and CIFAR-100. The compared methods include:
\begin{itemize}
\item ILM-VP~\cite{chen2023understanding} explores an iterative label mapping method to adaptively remap source labels to target classes.
\item LP (Linear Probing) adapts to downstream tasks by training a classifier head.
\item AutoVP~\cite{tsao2023autovp} achieves the optimal balance between image resizing scales and prompt frame sizes.
\item EVP~\cite{wu2024unleashing} shrinks the input image and pads learnable prompts around the borders to preserve original image information, while utilizing $L_2$ gradient normalization for stable optimization.
\item LoR-VP~\cite{jin2025lor} generates dense pixel-level prompts for each pixel via low-rank matrix multiplication.
\end{itemize}
As shown in Table~\ref{tab:vis}, LoRSP achieves excellent performance on both datasets, demonstrating effectiveness and generalization across CNN, transformer-based, and vision-language architectures. On Tiny-ImageNet, LoRSP achieves an average performance of 83.70\% across different backbones, outperforming LP by 2.94\% and the dense pixel-level prompting method LoR-VP by 1.22\%. A similar trend is observed on CIFAR-100, where LoRSP obtains the best average performance of 85.03\%. These consistent performance gains demonstrate that the proposed sparse visual prompts provide a more effective adaptation mechanism than existing visual prompting strategies, by suppressing redundant prompt responses and preserving task-relevant information.

\begin{table*}[htbp]
  \centering
  \caption{OOD generalization from ImageNet-1K to ImageNet variants using ViT-B/32 as the pre-trained backbone. ``Source'' denotes in-distribution accuracy on ImageNet-1K, while ``Target'' reports accuracy on OOD datasets.}
    \begin{tabular}{c|c|cccc|c}
    \toprule
    \multirow{2}[4]{*}{Method} & Source & \multicolumn{5}{c}{Target} \\
\cmidrule{2-7}          & \multicolumn{1}{l|}{ImageNet-1K} & \multicolumn{1}{l}{ImageNet-R} & \multicolumn{1}{l}{ImageNet-Sketch} & \multicolumn{1}{l}{ImageNet-A} & \multicolumn{1}{l|}{ImageNet-V2} & Avg \\
    \midrule
    *LP    & 77.69 & 32.06 & 34.18 & 19.29 & 64.57 & 37.53 \\
    *EVP  & 77.85 & 32.56 & 34.19 & 20.04 & 65.11 & 37.98 \\
    *LoR-VP & 77.89 & 32.23 & 34.02 & 19.96 & 65.15 & 37.84 \\
    \rowcolor[rgb]{ .906,  .902,  .902} LoRSP (Ours)   & \textbf{78.78} & \textbf{32.86} & \textbf{35.67} & \textbf{20.61} & \textbf{65.74} & \textbf{38.72} \\
    \bottomrule
    \end{tabular}%
  \label{tab:ood}%
\end{table*}%

\subsubsection{Performance on diverse downstream tasks}
To further assess the effectiveness of our proposed LoRSP, we conduct experiments on eight diverse downstream classification tasks. The tasks cover a wide range of visual domains: generic object recognition (Tiny-ImageNet, CIFAR-10/100), fine-grained classification (OxfordPets, Flowers102, Food101), remote sensing (EuroSAT), and texture recognition (DTD). We compare LoRSP with representative methods, including LP, VPT~\cite{jia2022visual}, AutoVP~\cite{tsao2023autovp}, and LoR-VP~\cite{jin2025lor}. We use ViT-B/32 as the pre-trained backbone for feature extraction. Table~\ref{tab:tasks} shows that LoRSP achieves the best average accuracy of 90.54\% across the eight tasks. Compared with LP, LoRSP improves the average accuracy by +2.51\%, and further outperforms LoR-VP and VPT by 1.24\% and 2.19\%, respectively. Moreover, LoRSP exhibits remarkable performance on specialized domains. In particular, on texture classification (DTD), it improves LP by 6.88\% (76.71\% vs. 69.83\%), while on remote sensing (EuroSAT), it achieves an accuracy of 97.09\%. These results indicate the effectiveness of LoRSP for downstream adaptation by combining low-rank prompt factorization and SNN to generate pixel-level sparse prompts for each instance. 

\subsubsection{Out-of-distribution generalization}
We evaluate the out-of-distribution (OOD) robustness of LoRSP using ViT-B/32 as the backbone. Following the previous works~\cite{jin2025lor,khattak2023maple}, the model is trained on ImageNet-1K and directly tested on four challenging OOD benchmarks: ImageNet-R, ImageNet-Sketch, ImageNet-A, and ImageNet-V2 (Table~\ref{tab:ood}). As shown in Table~\ref{tab:ood}, LoRSP consistently achieves the best performance across all OOD datasets, improving the average OOD accuracy from 37.53\% (LP), 37.98\% (EVP), and 37.84\% (LoR-VP) to 38.72\%. Meanwhile, LoRSP maintains competitive in-distribution performance on ImageNet-1K (78.78\%). These results indicate that LoRSP improves robustness to distribution shifts while preserving competitive performance on the source domain.


\subsection{Ablation studies}
\subsubsection{Effectiveness of different components}
To validate the effectiveness of the key components of LoRSP, we conduct ablation experiments on the low-rank prompt factorization and SNN-based sparse prompt learning. LP serves as the baseline, where a frozen pre-trained backbone is adapted to the target task using only a linear classification head. Based on this baseline, LPF introduces the low-rank prompt factorization module, where the prompt factors are averaged to obtain dense instance-specific visual prompts. The LoRSP is the proposed method that integrates low-rank prompt factorization and SNN-based sparse prompt learning modules. We evaluate these variants on multiple large-scale pre-trained vision backbones, and report the results in Table~\ref{ablat}. LPF consistently improves the baseline across different backbones and both datasets. On Tiny-ImageNet, LPF improves accuracy by +1.51\% on ResNet-50, +2.29\% on ViT-B/32, and +5.33\% on CLIP. Similar trends are observed on the CIFAR-100 dataset, where LPF achieves gains of 2.66\% on ResNet-50 and 1.84\% on ViT-B/32. These results suggest that constructing a series of low-rank prompt factors to capture distinct semantic subspaces is effective for downstream adaptation.
Furthermore, incorporating the SNN-based sparse prompt learning module further improves performance (e.g., +1.26\% on Swin-B), indicating that the SNN can effectively accumulate contextual information across prompt factors and filter out redundant and task-irrelevant prompt components, thereby reducing information interference and improving downstream adaptation.

\begin{table}[htbp]
  \centering
  \caption{Ablation experiments of different components on various larger-scale pre-trained vision backbones.}
    \resizebox{0.5\textwidth}{!}{
    \begin{tabular}{c|c|rrr}
    \toprule
    Dataset & Vision Backbone & \multicolumn{1}{c}{Baseline} & \multicolumn{1}{c}{LPF} & \multicolumn{1}{c}{LoRSP (Ours)} \\
    \midrule
    \multirow{6}[4]{*}{Tiny-ImageNet} & ResNet-50 & \multicolumn{1}{c}{76.52} & \multicolumn{1}{c}{78.03} & \multicolumn{1}{c}{\textbf{78.27}} \\
          & ViT-B/32 & \multicolumn{1}{c}{83.95} & \multicolumn{1}{c}{86.24} & \multicolumn{1}{c}{\textbf{86.93}} \\
          & ViT-B/16 & \multicolumn{1}{c}{88.86} & \multicolumn{1}{c}{90.08} & \multicolumn{1}{c}{\textbf{90.78}} \\
          & Swin-B & \multicolumn{1}{c}{86.54} &  87.20     & \multicolumn{1}{c}{\textbf{88.46}}\\
          & CLIP  & \multicolumn{1}{c}{67.93} & \multicolumn{1}{c}{73.26} & \multicolumn{1}{c}{\textbf{74.06}} \\         
    \midrule
    \multirow{6}[4]{*}{ CIFAR-100} & ResNet-50 & \multicolumn{1}{c}{73.18} & \multicolumn{1}{c}{75.84} & \multicolumn{1}{c}{\textbf{76.34}} \\
          & ViT-B/32 & \multicolumn{1}{c}{86.21} & \multicolumn{1}{c}{88.05} & \multicolumn{1}{c}{{\textbf{88.69}}} \\
          & ViT-B/16 & \multicolumn{1}{c}{88.90} & \multicolumn{1}{c}{89.13} & \multicolumn{1}{c}{\textbf{89.62}} \\
          & Swin-B & \multicolumn{1}{c}{88.52} & 89.49      & \multicolumn{1}{c}{\textbf{90.16}} \\
          & CLIP  & \multicolumn{1}{c}{78.15} & \multicolumn{1}{c}{79.61} & \multicolumn{1}{c}{\textbf{80.36}} \\
    \bottomrule
    \end{tabular}%
  \label{ablat}%
  }
\end{table}%

\subsubsection{Comparison of sparsification strategies}
To validate the effectiveness of our SNN-based sparse prompt learning, we compare it with the representative sparsification methods, including Top-$k$ selection~\cite{jayakumar2020top} and random dropout~\cite{srivastava2014dropout}, under different sparsity ratios (10\%, 20\%, 50\%, 80\%). For the Top-$k$ strategy, we retain the $k$ highest-response prompts and set the remaining prompts to zero. Random dropout stochastically masks prompts to the target activation ratio. The results are reported in Table~\ref{re_sparse}. Our LoRSP consistently outperforms Top-$k$ selection and random dropout across all sparsity ratios and backbones. Unlike conventional sparsification methods, which impose sparse constraints independently of prompt learning, our LoRSP performs instance-specific sparse selection within a unified prompt learning framework. This advantage mainly comes from two aspects. First, Top-$k$ selection retains a fixed number of prompts for each instance, making it difficult to adapt the sparsity to different inputs, while random dropout masks prompts stochastically and may suppress task-relevant cues. Second, these methods do not exploit the contextual information across prompt factors for adaptive sparse selection. In contrast, the proposed SNN-based sparse prompt learning accumulates contextual information across prompt factors and produces instance-specific sparse visual prompts. As a result, LoRSP suppresses redundant and task-irrelevant prompt responses while preserving informative and complementary visual cues, leading to more effective downstream adaptation.

\begin{table}[htbp]
  \centering
  \caption{Impact of different sparsity strategies. We compare our LoRSP with Top-$k$ sparsification and random dropout under different sparsity ratios (10\%, 20\%, 50\%, 80\%) on two backbones (ViT-B/32 and ResNet-50).}
  \resizebox{0.5\textwidth}{!}{
  \setlength{\tabcolsep}{10pt}
    \begin{tabular}{c|c|c|c|c}
    \toprule
    Network & Sparse & Top-k & Dropout & LoRSP (Ours) \\ 
    \midrule
    \multirow{4}[2]{*}{ViT-B/32} & 10\%  & 86.23  & 86.25  & \multirow{4}[2]{*}{\textbf{86.93}} \\
          & 20\%  & 86.21  & 86.37  &  \\
          & 50\%  & 86.40  & 86.44  &  \\
          & 80\%  & 86.38  & 86.40  &  \\
    \midrule
    \multicolumn{1}{c|}{\multirow{4}[2]{*}{ResNet-50}} & 10\%  & 77.18 & 77.42 & \multirow{4}[2]{*}{\textbf{78.27}} \\
          & 20\%  & 77.47 & 77.46 &  \\
          & 50\%  & 77.27 & 77.32 &  \\
          & 80\%  & 77.17 & 77.57 &  \\
    \bottomrule
    \end{tabular}%
    }
  \label{re_sparse}%
\end{table}%

\begin{figure}[htbp]
  \centering
  \includegraphics[width=1\linewidth]{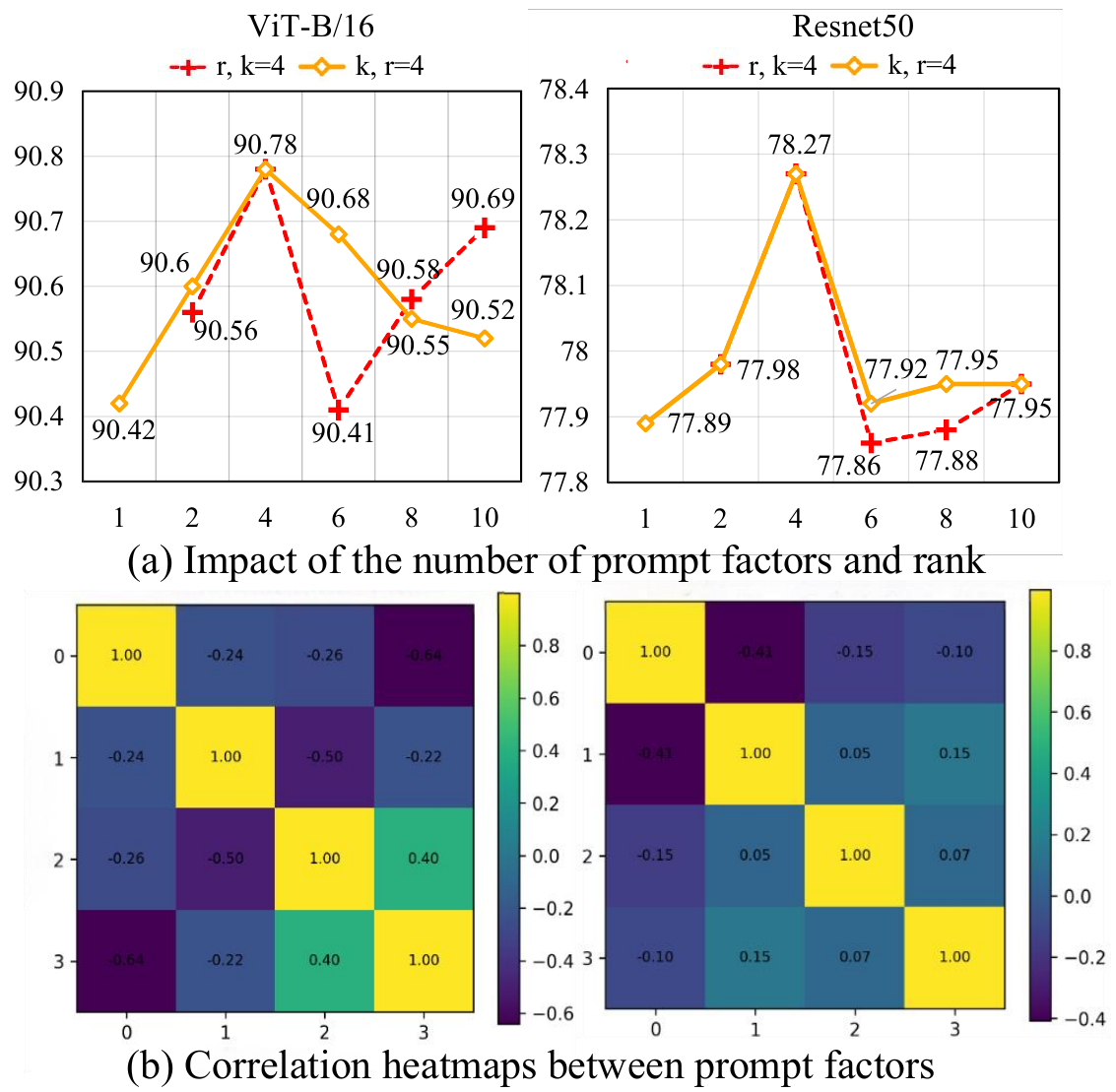}
  \caption{ Impact of the number of prompt factors and rank in LoRSP and correlation heatmaps between prompt factors.} 
\label{e_mdb}
\end{figure}

\begin{figure}[htbp]
  \centering
  \includegraphics[width=1\linewidth]{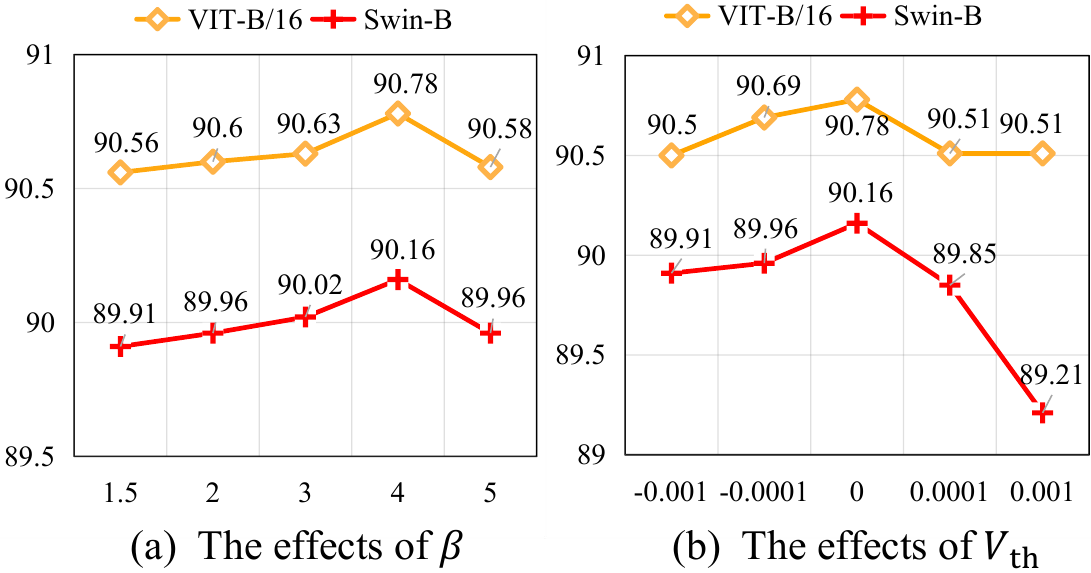}
  \caption{ Effects of the $V_{th}$ and $\beta$ in LoRSP on two backbones (ViT-B/16 and Swin-B). } 
\label{e_ssa}
\end{figure}

\begin{figure*}[htbp]
  \centering
  \includegraphics[width=0.9\linewidth]{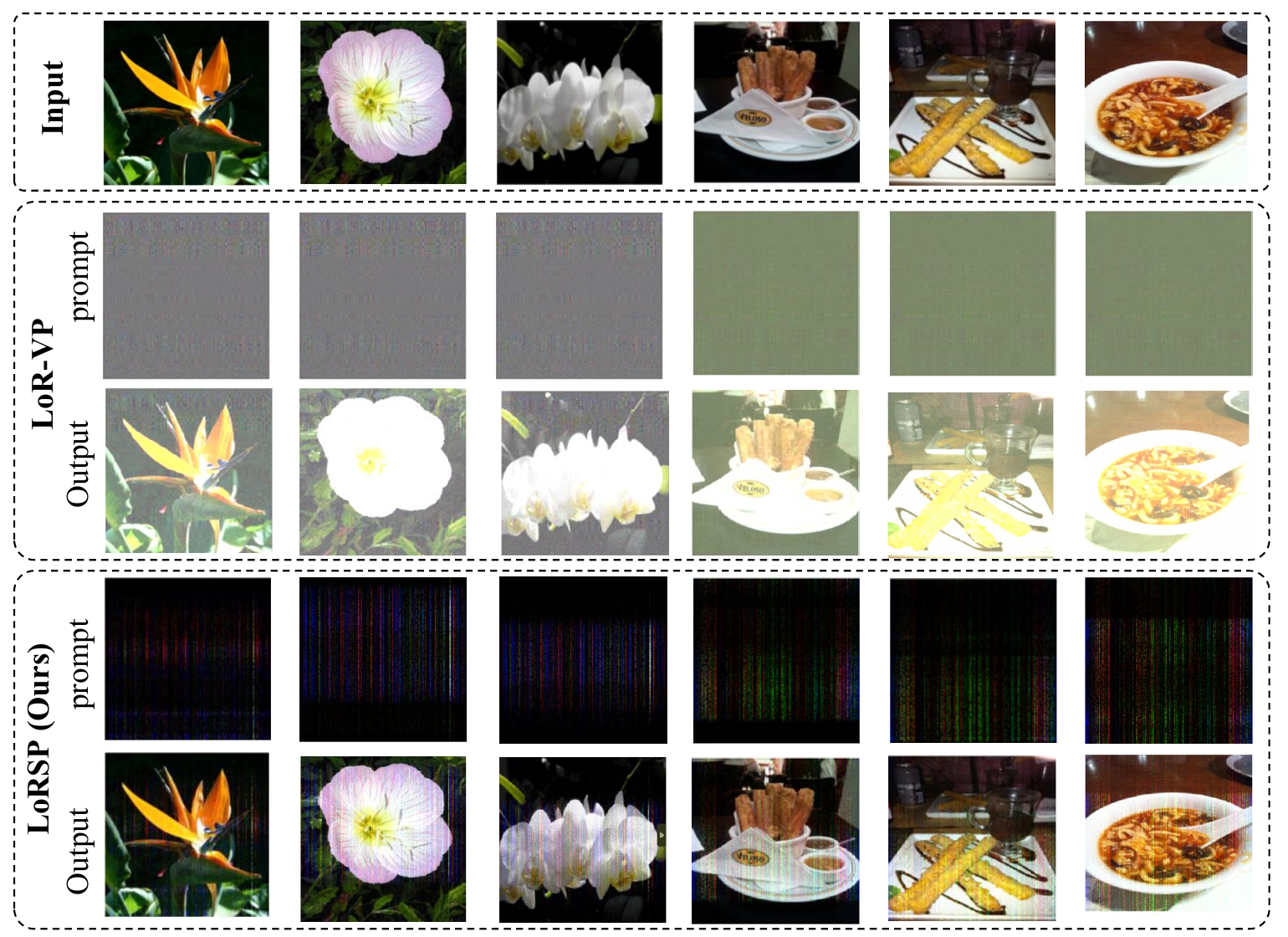}
  \caption{Visualization of learned visual prompts. We compare LoR-VP and our LoRSP by visualizing the input images, the learned visual prompts, and the corresponding prompted images. 
  } 
\label{view}
\end{figure*}

\subsection{Hyperparameters analysis}
\subsubsection{Effect of the  prompt number of factors and rank}
We analyze the sensitivity of LoRSP to two key hyperparameters: the number of prompt factors $k$ and the rank $r$.  As illustrated in Fig.~\ref{e_mdb} (a), we vary $r\in\{2,4,6,8,10\}$ with a fixed $k=4$ (red line), and vary $k\in\{1,2,4,6,8,10\}$ with a fixed $r=4$ (yellow line). Experiments are conducted on Tiny-ImageNet with two representative backbones, ViT-B/16 and ResNet-50. We observe superior performance at $k=4$ and $r=4$ for both backbones. 
When using a single prompt factor ($k=1$), performance consistently drops compared to multiple prompt factors on both backbones. Increasing the number of prompt factors ( $k>4$) or the rank ($r>4$) leads to a slight drop in performance. These results indicate that a larger rank or a greater number of prompt factors can introduce redundant prompt components, while small $k$ or $r$ limit the capacity to capture instance-specific cues. Therefore, we utilize $k=4$ and $r=4$ as the default setting. Notably, for the OOD setting, we set $k=2$ and $r=2$, which encourages the model to focus on more transferable cues while suppressing task-irrelevant prompt components, improving generalization under distribution shifts. 

In addition, to verify the complementarity of the learned prompt factors, we visualize heatmaps of pairwise correlations between prompt factors in Fig.~\ref{e_mdb} (b). The correlation heatmaps show that the prompt factors are weakly correlated and complementary across subspaces. These results verify that LoRSP effectively learns diverse and complementary prompt factors, allowing each instance to be represented in distinct semantic subspaces and thus leading to better downstream adaptation.

\subsubsection{Effect of the $V_{th}$ and $\beta$ in LoRSP}
We further analyze the effect of $\beta$ and $V_{th}$ on ViT-B/16 (Tiny-ImageNet) and Swin-B (CIFAR-100). The corresponding results are shown in Fig.~\ref{e_ssa}. Here, the leak factor $\beta$ denotes the exponential decay of the membrane potential over time. $V_{th}$ denotes the firing threshold and controls the degree of sparsification. The optimal performance is obtained at $\beta=4$ and $V_{th}=0$. We observe consistent performance gains when increasing $\beta$ from 1.5 to 4. This suggests that an appropriate leak factor helps the model effectively accumulate contextual information from multiple prompt factors. In contrast, a large $\beta$ weakens the contribution of subsequent prompt factors during the accumulation process, thereby limiting downstream adaptation.
Similarly, increasing $V_{th}$ generally promotes sparser activations. However, a high threshold may filter out too many informative responses, limiting the model’s ability to refine task-specific visual cues and leading to performance degradation.

\subsection{Visualization of learned visual prompts}

In Fig.~\ref{view}, we visualize the prompts produced by LoR-VP~\cite{jin2025lor} and our proposed LoRSP. For each example, we present the input image, the learned visual prompts, and the corresponding prompted image. As shown in Fig.~\ref{view}, LoR-VP generates shared and dense visual prompts for each pixel of the input image. Such dense prompts are instance-invariant, limiting their ability to adapt to diverse image samples. Moreover, they may introduce redundant perturbations and disrupt the underlying spatial structure, limiting generalization on downstream tasks. In contrast, LoRSP produces sparse and instance-specific visual prompts. From the Fig.~\ref{view}, we observe that the prompts are mainly located on the foreground target regions (e.g., flower centers, and food areas), while the background regions are largely suppressed. These results indicate that our LoRSP filters out redundant prompt responses while preserving salient task-relevant cues, leading to more effective downstream adaptation.

\subsection{Efficiency Analysis}
We evaluate the parameter efficiency of LoRSP by reporting the number of trainable visual prompt parameters. Given a resized input of resolution \(c \times L \times L\), the number of trainable parameters for visual prompts in LoRSP is given by:
\begin{align}
&N_{\mathrm{LoRSP}} = L \times r \times (c+k),\\
&N_{\mathrm{AB}} = 2\times L \times c \times r \times k,\\
&N_{\mathrm{LoRSP}} < N_{\mathrm{AB}}
\quad \Leftrightarrow \quad
k \geq 1,
\end{align}
where \(c=3\) denotes the number of input channels and \(L\) denotes the width of the resized image. Here, \(k\) denotes the number of prompt-factor groups and \(r\) is the low-rank decomposition rank. The grouped low-rank design of LoRSP reduces parameter redundancy relative to the ungrouped formulation $N_{\mathrm{AB}}$.
For downstream adaptation across various vision backbones, we use \(k=4\) and \(r=4\), requiring only 6.27K trainable parameters. This is substantially smaller than ILM-VP (147K) and AutoVP (74K). For OOD adaptation, we further reduce the setting to \(k=2\) and \(r=2\), which requires only 2.24K trainable parameters. These results demonstrate that LoRSP retains strong adaptation capability with fewer trainable parameters.

\section{Conclusion}
In this paper, we propose \textbf{Lo}w-\textbf{R}ank visual \textbf{S}pike \textbf{P}rompting (LoRSP), a parameter-efficient visual prompting framework for adapting frozen pre-trained vision backbones to downstream tasks. To address the redundancy and limited generalization of existing dense pixel-level prompts, LoRSP integrates low-rank prompt factorization with a spiking neural network to generate instance-specific sparse visual prompts. Specifically, we construct a series of low-rank prompt factors to capture distinct semantic subspaces for each input image, and further employ an SNN-based sparse prompt learning to adaptively aggregate these prompt factors, producing instance-specific sparse visual prompts. In this way, LoRSP preserves the low-rank constraint while enabling instance-specific selective prompting. Extensive experiments across diverse backbones and datasets demonstrate that LoRSP achieves competitive performance with a small number of tunable parameters, providing an effective and efficient method compared to existing visual prompting methods.

\bibliographystyle{IEEEtran}
\bibliography{reference}
\end{document}